\documentclass[times, 10pt,twocolumn]{article} 
\usepackage{latex8}
\usepackage{times}
\usepackage{amsmath,amssymb}
\usepackage{graphicx}

\begin{document}

\title{Atmospheric Turbulence Correction via Variational Deep Diffusion}

\author{Xijun Wang$^1$, Santiago López-Tapia$^2$, Aggelos K. Katsaggelos$^2$\\
$^1$Dept. of Computer Science, Northwestern University, Evanston, IL, USA\\
$^2$Dept. of Electrical and Computer Engineering, Northwestern University, Evanston, IL, USA\\ xijunwang2022@u.northwestern.edu
}

\maketitle
\thispagestyle{empty}

\begin{abstract}
    Atmospheric Turbulence (AT) correction is a challenging restoration task as it consists of two distortions: geometric distortion and spatially variant blur. Diffusion models have shown impressive accomplishments in photo-realistic image synthesis and beyond. In this paper,  we propose a novel deep conditional diffusion model under a variational inference framework to solve the AT correction problem. We use this framework to improve performance by learning latent prior information from the input and degradation processes. We use the learned information to further condition the diffusion model. Experiments are conducted in a comprehensive synthetic AT dataset. We show that the proposed framework achieves good quantitative and qualitative results.
\end{abstract}
\let\thefootnote\relax\footnote{© 20XX IEEE.  Personal use of this material is permitted.  Permission from IEEE must be obtained for all other uses, in any current or future media, including reprinting/republishing this material for advertising or promotional purposes, creating new collective works, for resale or redistribution to servers or lists, or reuse of any copyrighted component of this work in other works.

This research is based upon work supported in part by the Office of the Director of National Intelligence (ODNI), Intelligence Advanced Research Projects Activity (IARPA), via [2022-21102100007]. The views and conclusions contained herein are those of the authors and should not be interpreted as necessarily representing the official policies, either expressed or implied, of ODNI, IARPA, or the U.S. Government. The U.S. Government is authorized to reproduce and distribute reprints for governmental purposes notwithstanding any copyright annotation therein.}
\Section{Introduction}
Atmospheric Turbulence is an issue in real-life long-range imaging caused by slight perturbations in atmospheric conditions (e.g., temperature), and it can cause severe blur and perceptual degradation. This, in turn, could severely effect performance in the subsequent downstream vision tasks, such as detection, recognition, and so on. Unlike other imaging inverse problems, atmospheric turbulence degradation contains a mixture of geometrical distortion, spatially variant blur, and noise, which makes AT more challenging to mitigate. 

\vspace{0.2cm}

Earlier works in AT correction mainly focus on optics and lucky imaging algorithms \cite{zhu2012removing}.  These algorithms are often computationally expensive. In recent years, with the development of deep-learning (DL) algorithms for solving various inverse problems \cite{wang2020deep}, some works have proposed DL-based AT removal methods \cite{mao2022single,rai2022removing}. With \cite{mao2021accelerating} proposing a fast AT simulation algorithm, large-scale data-driven DL training for AT correction becomes possible \cite{mao2022single}. In this paper, we also adopt the simulation method in \cite{mao2021accelerating} to construct our training and testing datasets.

Recently, deep diffusion models have been proposed and developed for image generation \cite{ho2020denoising}. As a likelihood-based algorithm, it is more stable during training than generative adversarial networks (GAN) and does not suffer from mode collapse. Diffusion models have shown significant success in various vision problems, like image synthesis \cite{rombach2022high} and super-resolution \cite{rombach2022high, saharia2022image}. In a very recent work,  diffusion models are used for the atmospheric turbulence restoration of faces \cite{nair2023ddpm}. However, no published work has addressed the use of diffusion models in generic scenes AT correction. In this paper,  we propose a diffusion model to remove atmospheric turbulence in generic scenes, producing results with great visual quality.  In addition, we refer to a variational inference image restoration framework \cite{soh2022variational} to learn the latent features related to task-specific prior information from the input and the degradation process; we then inject this learned knowledge as a condition into the diffusion models. Therefore, the diffusion model is trained to adjust its behavior according to both the input degraded image and the task-specific prior information, which further enhances its performance.  
\vspace{-0.01cm}

In summary, our main contributions are: 
1) We are the first to use diffusion models to solve the AT correction problem in generic scenes. 
2) We propose to include a variational inference framework to provide a task-specific condition to the diffusion models. 
3) We show that our proposed AT \textbf{var}iational deep \textbf{diff}usion (AT-VarDiff) model generates results with outstanding visual quality evaluated both quantitatively and qualitatively.
\Section{Method}
\label{sec: method}

In this section, we present the proposed AT-VarDiff model. We explain our model's conditional denoising diffusion process in Section \ref{subsec:cdg}. In Section \ref{subsec:vif} we introduce the variational framework used to obtain the condition encoding task-specific information. Finally, in Section \ref{subsec:inference} we illustrate the inference during testing.
\SubSection{Conditional Diffusion Model}
\label{subsec:cdg}

\begin{figure}[h]
\centering
\centerline{\includegraphics[width=8 cm]{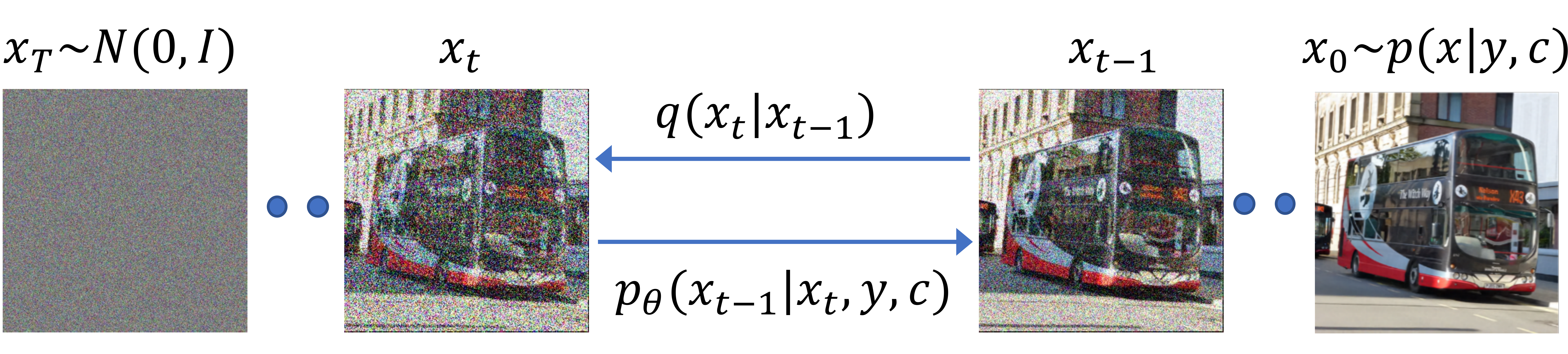}}
\caption{Conditional denoising diffusion process.}
\label{fig:Cond_diff}
\end{figure}

Our training dataset contains \(N\) image pairs \(\{y_i, x_i\}_{i=1}^N\), where \(y_i\) represents the AT degraded image and \(x_i\) the corresponding ground-truth image. As shown in Figure \ref{fig:Cond_diff}, our model aims at learning the data distribution \(p(x|y, c)\) by a stochastic iterative refinement process, which maps the input degraded image \(y\) and the learned latent prior information \(c\)  to the ground-truth image \(x\).  The forward/diffusion process (from right to left) gradually adds Gaussian noise, denoted by \(q(x_t|x_{t-1})\).  Our goal is to reverse the diffusion process (from left to right) by gradually recovering the image from the input Gaussian noise with conditions, which corresponds to learning the reverse process of a fixed Markov Chain of length \(T\) conditioned on \textit{y} and \(c\). More specifically, starting from a pure Gaussian noise image \(x_T\sim \mathcal{N}(\textbf{\textit{0}}, \textbf{\textit{I}})\), the model learns the conditional transition distribution \(p_\theta(x_{t-1}|x_{t}, y, c)\) and iteratively denoises the image for \(T\) steps, generating the target image \(x_0\) in the end, such that \(x_0 \sim p(x|y,c)\). 

\begin{figure}[h]
\centering
\centerline{\includegraphics[width=7.2 cm]{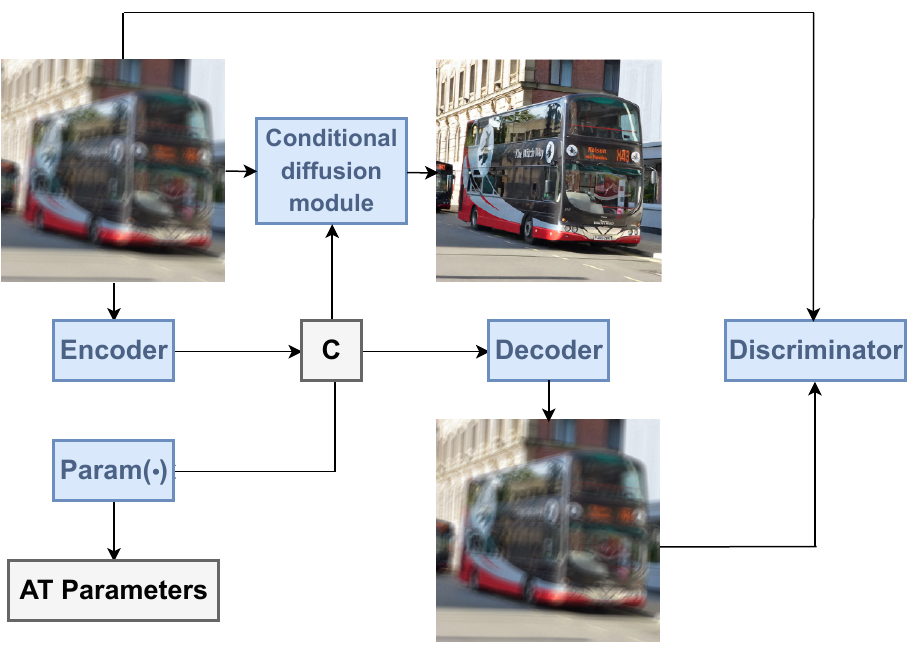}}
\caption{Training framework of AT-VarDiff model.}
\label{fig:model_train}
\vspace{-0.3cm}
\end{figure}

The overall training framework of the AT-VarDiff model is shown in Figure \ref{fig:model_train}. Following the model design in denoising diffusion probabilistic model (DDPM) \cite{ho2020denoising}, the architecture of our conditional diffusion module is a U-Net \cite{ronneberger2015u} based on a wide ResNet \cite{zagoruyko2016wide}, denoted as \(\epsilon_\theta\). Training is performed by optimizing the usual variational bound on the negative log-likelihood, and the corresponding objective function can be simplified to \cite{ho2020denoising, rombach2022high}: 
\begin{equation}
\label{eq:loss_diff}
\mathcal{L}_{diff} = \mathbb{E}_{x, y, c, \varepsilon, t}[\|\epsilon - \epsilon_\theta(x_t, t, y, c) \|_2^2], 
\end{equation}
with \(t\) is uniformly sampled from \(\{1, ..., T\}\). According to Equation 1,  \(\epsilon_\theta\) takes as input the noisy target image \(x_t\), time step \(t\), the AT degraded image \(y\), and the learned latent prior information \(c\)  to provide an estimate of the noise \(\epsilon\). The details of obtaining \(c\) are discussed in the following Section.

\SubSection{Variational Inference Framework}
\label{subsec:vif}
The current conditional diffusion models used for solving image restoration tasks like super-resolution \cite{saharia2022image, rombach2022high}, and face AT correction \cite{nair2023ddpm} only use the input degraded image as the condition. No work in the literature has employed any other task-specific prior information or domain-knowledge  to further enhance the conditioning progress. According to \cite{soh2022variational}, providing additional information can be interpreted as dividing a complex distribution into simpler sub-distributions that will eventually make network training easier and the results more accurate, since the number of possible solutions would be reduced. In this paper, we propose to use a variational inference framework to extract the latent task-specific prior information from the input and the degradation process and use the extracted feature as an additional condition to guide the diffusion model.

As shown in Figure \ref{fig:model_train}, we refer to a variational autoencoder (VAE) based framework \cite{soh2022variational} to learn the latent feature \(c\) from the input degraded image \(y\) and the AT degradation parameters.  To achieve this goal, the objective we use here contains three parts: the VAE loss, the adversarial loss, and the AT degradation parameters' loss. 

The VAE loss contains the fidelity term and the reconstruction term, that is, 
\begin{equation}
\label{eq:loss_vae}
\mathcal{L}_{vae} = D_{KL}(q_{e_\psi}(c|y)||p(c)) + ||y - \hat{y}||_2^2. 
\end{equation}
\begin{figure}[h]
\centering
\centerline{\includegraphics[width=5.6 cm]{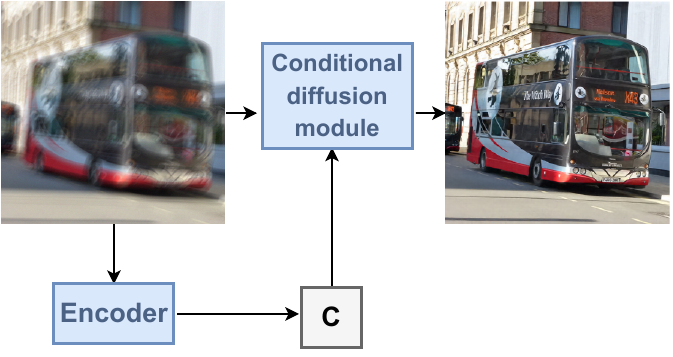}}
\setlength{\abovecaptionskip}{-12pt}
\caption{Testing framework of AT-VarDiff model.}
\label{fig:model_test}
\vspace{-0.5cm}
\end{figure}
The first term is the fidelity term; it measures the fidelity of \(c\) extracted from the encoder \(e_\psi\), whose input is the degraded image \(y\). It is represented as the KL divergence of the approximate posterior \(q_{e_\psi}(c|y)\) from the prior \(p(c)\). We select the prior \(p(c)\) as a standard Gaussian distribution.  The second term is the reconstruction term, and we adopt the pixel-wise mean squared error (MSE) distance between the input degraded image \(y\) and the output \(\hat{y}\) of the decoder \(d_\varphi\). In addition, we utilize a GAN \cite{goodfellow2020generative} to better learn the input degraded image distribution, in which an additional discriminator is jointly trained to discriminate the generated \(\hat{y}\) and the true degraded image \(y\). Therefore, we also include an adversarial loss
\begin{equation}
\label{eq:loss_adv}
\mathcal{L}_{adv} = -log(D(\hat{y})),
\end{equation}
and the corresponding loss for the discriminator \(D\) is
\begin{equation}
\label{eq:loss_dis}
\mathcal{L}_{disc} = -log(D(y)) - log(1-D(\hat{y})).
\end{equation}
Finally, we would like the latent feature \(c\) to contain knowledge from the AT degradation process. Therefore, we add a degradation loss defined as:
\begin{equation}
\label{eq:loss_degrad}
\mathcal{L}_{degrad} = ||\phi_{at} - \hat{\phi_{at}}||_2^2, 
\end{equation}
where \(\phi_{at}\) represents the ground-truth AT degradation parameters from the pre-trained AT simulator \cite{mao2021accelerating}. \(\hat{\phi_{at}} = Param(c)\) represents the estimated AT degradation parameters, and is the output of a small network (parameter estimation module) \(Param(\cdot)\) taking \(c\) as input, as shown in Figure \ref{fig:model_train}. 

Therefore, the final objective used for training our AT-VarDiff model is defined as
\begin{equation}
\label{eq:loss_final}
\mathcal{L} = \mathcal{L}_{diff} + \lambda_{1}\mathcal{L}_{vae} + \lambda_{2}\mathcal{L}_{adv} + \lambda_{3}\mathcal{L}_{degrad},
\end{equation}
where \(\lambda_1, \lambda_2, \lambda_3\) are hyper-parameters. 
\SubSection{Inference}
\vspace{-0.2cm}
\label{subsec:inference}
The testing framework of our model is shown in Figure \ref{fig:model_test}. During testing, we followed the DDPM's denoising sampling procedure (Algorithm 2 in \cite{ho2020denoising}) conditioned on the input degraded image \(y\) and the learned task-specific latent feature \(c\) to generate the output restored image. During both training and testing, we perform the conditioning via concatenation, and we set \(T = 1000\) for all the experiments.
\begin{table}[htbp]
\caption{LPIPS \& FID metrics comparison on simple-DDPM, AT-VarDiff, and AT-DDPM \cite{nair2023ddpm}.}
\begin{center}
\small
\resizebox{0.9\columnwidth}{!}{
\begin{tabular}{cccc}
 \hline
 & \textbf{AT-DDPM \cite{nair2023ddpm}} & \textbf{Simple-DDPM} & \textbf{AT-VarDiff (Ours)}\\ 
 \hline\hline
\textbf{LPIPS \(\downarrow\)} & 0.2150 & 0.1923 & \textbf{0.1094} \\ 
 \hline
\textbf{FID \(\downarrow\)} & 80.05 & 60.87 & \textbf{32.69}  \\ 
 \hline
\end{tabular}
}
\end{center}
\label{tab:table_metric}
\vspace{-0.6cm}
\end{table}
\vspace{-0.2cm}
\Section{Experiments}
\vspace{-0.2cm}
\subsection{Experimental Settings}
We use the AT simulator in \cite{mao2021accelerating} to simulate the effect of AT on the REDS dataset \cite{nah2019ntire}, and the hyper-parameter of the simulator (\(D/r_0\)) is chosen randomly in the range [0.5, 2.0].  Our synthetic training dataset has one million AT degraded and clean image pairs. We use another 2500 synthetic AT degraded images as the testing dataset.

For our encoder module, we use 5 2D-convolution (2D-conv) layers with ReLU activation and one down-sampling layer after the first conv layer. For the decoder module, we use 5 2D-conv layers with ReLU activation and one up-sampling layer after the first conv layer. For our parameter estimation module, we simply use 2 2D-conv layers with LeakyReLU activation. The discriminator is formed by 11 2D-conv layers with LeakyReLU activation and spectral normalization \cite{miyato2018spectral}.  

During training, we augment the training data by random cropping (\(160\times\)160), random vertical and horizontal flips, and random transposing. We train our model for 200 epochs with 1500 iterations per epoch, and we set the batch size to 16. We use the Adam optimizer \cite{kingma2014adam} with a weight decay of 0, and we set the initial learning rate to 1\(e-4\) and gradually reduced it to 5\(e-6\) during training utilizing the cosine annealing schedule \cite{loshchilov2016sgdr}. The hyper-parameters \(\lambda_1, \lambda_2\), and \(\lambda_3\) used in our final training objective (Equation \ref{eq:loss_final}) are set to 0.1, 0.1, and 0.5, respectively.
\begin{figure}[h]
\centering
\centerline{\includegraphics[width=7.3 cm]{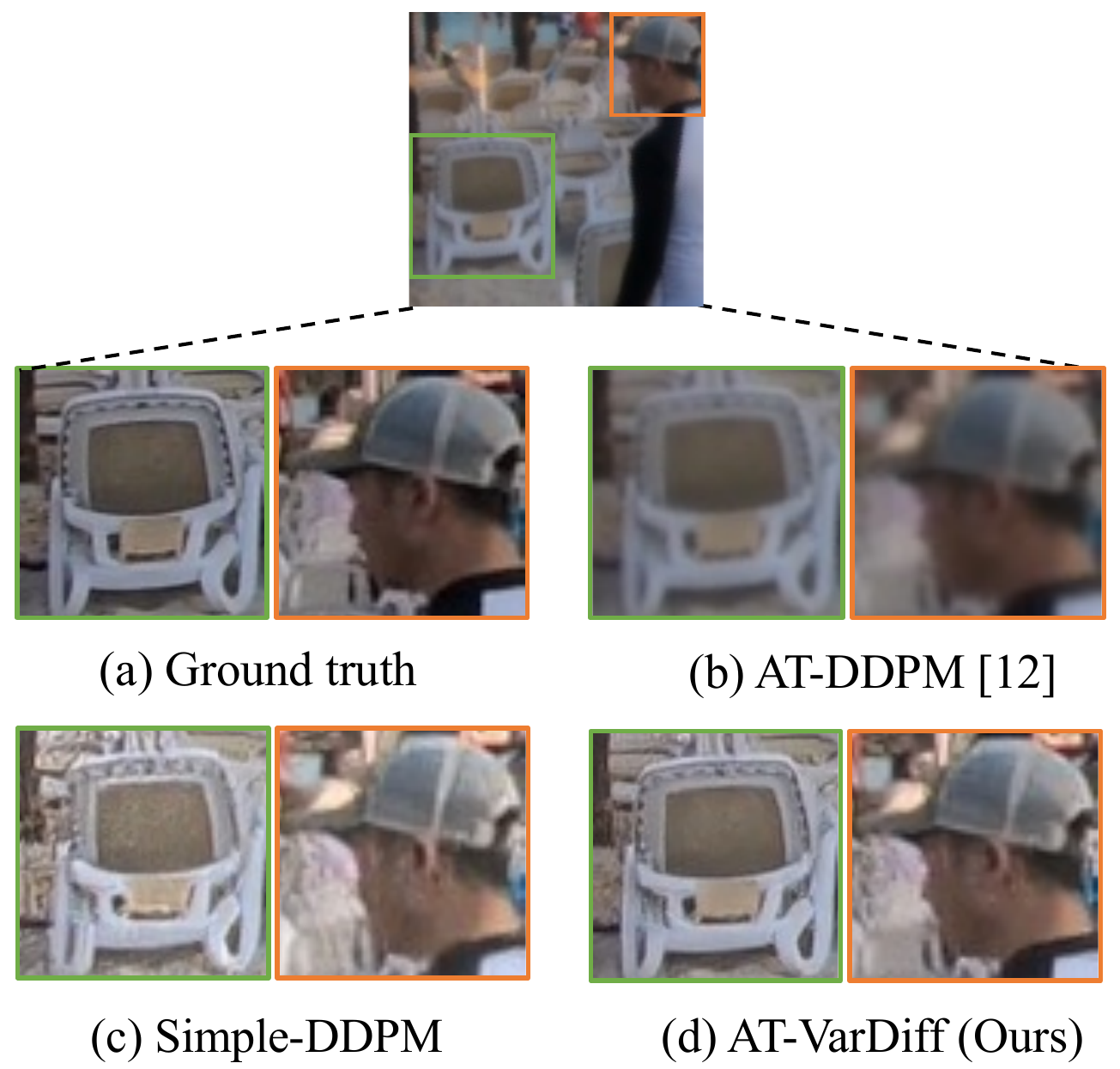}}
\caption{Visual comparisons of AT-DDPM \cite{nair2023ddpm}, simple-DDPM, and AT-VarDiff.}
\label{fig:result_img}
\vspace{-0.6cm}
\end{figure}
\subsection{Results}
\vspace{-0.2cm}
To evaluate our model, we use the Fréchet Inception Distance 
(FID) \cite{heusel2017gans} and the Learned Perceptual Image Patch Similarity (LPIPS) \cite{zhang2018unreasonable} metrics, which are measures of similarity between two sets of images. These two metrics are shown to correlate well with the human judgment of visual quality. In Table \ref{tab:table_metric}, we show the quantitative results of our proposed AT-VarDiff model and compare it to using a pure conditional DDPM-based diffusion model like the approach used in \cite{nair2023ddpm}, i.e., this simple-DDPM model is only built with the conditional diffusion module and is only conditioned on the input degraded image \(y\). We can see that our AT-VarDiff model improves on both metrics, demonstrating the effectiveness of our proposed variational conditional diffusion framework.  We also compare with the pre-trained AT-DDPM model from \cite{nair2023ddpm} in the table. As can be seen in Figure \ref{fig:result_img}, our proposed approach achieves much better visual clarity, far fewer artifacts, and higher quality.
\Section{Conclusions}
\vspace{-0.2cm}
In this paper, we propose the variational deep diffusion model AT-VarDiff to restore images degraded by atmospheric turbulence. We propose to use the diffusion process to remove AT in generic scenes, and we use a variational inference framework to extract the latent task-specific prior information from the input and the AT degradation. We further inject extracted features as an additional condition to guide the diffusion model. We show that the proposed method achieves good results and outstanding visual quality, outperforming the current state-of-art. In the future, we will use more advanced diffusion techniques to further enhance the performance.
\vspace{-0.1cm}
\nocite{ex1,ex2}
\bibliographystyle{latex8}
\bibliography{latex8}

\end{document}